\UseRawInputEncoding
\documentclass[letterpaper, 10 pt, conference]{ieeeconf}  % Comment this line out if you need a4paper

\IEEEoverridecommandlockouts                              % This command is only needed if 
\usepackage{cite}
\usepackage{graphicx}
\usepackage{amsmath} % assumes amsmath package installed
\usepackage{amssymb}  % assumes amsmath package installed
\usepackage{xcolor}
\usepackage[hidelinks]{hyperref}
\usepackage{multirow}
\usepackage{makecell}
\usepackage{wrapfig}
\usepackage{balance}

\title{\Huge Mel-spectrogram features for acoustic\\ vehicle detection and speed estimation}
\author{Nikola Bulatovi\'c and Slobodan Djukanovi\'c, \it{Member, IEEE}% <-this % stops a space
\thanks{Nikola Bulatovi\'c and Slobodan Djukanovi\'c are with the Faculty of Electrical Engineering, University of Montenegro, Podgorica, Montenegro (e-mail: \{nbulatovic, slobdj\}@ucg.ac.me).}%
}
\usepackage{eso-pic}
\newcommand\AtPageUpperMyright[1]{\AtPageUpperLeft{%
		\put(\LenToUnit{0.5\paperwidth},\LenToUnit{-1cm}){%
			\parbox{0.5\textwidth}{\raggedleft\fontsize{9}{11}\selectfont #1}}%
}}%
\newcommand{\conf}[1]{%
	\AddToShipoutPictureBG*{%
		\AtPageUpperMyright{#1}
	}
}
\begin{document}
\pubid{\makebox[\columnwidth]{978-1-6654-2127-0/22/\$31.00 \copyright 2022 IEEE \hfill} \hspace{\columnsep}\makebox[\columnwidth]{ }}

\maketitle
\conf{26th International Conference on Information Technology (IT) \v{Z}abljak, 16 -- 19 February, 2022}
\pagestyle{empty}

%%%%%%%%%%%%%%%%%%%%%%%%%%%%%%%%%%%%%%%%%%%%%%%%%%%%%%%%%%%%%%%%%%%%%%%%%%%%%%%%

\begin{abstract}

The paper addresses acoustic vehicle detection and speed estimation from single sensor measurements. We predict the vehicle's pass-by instant by minimizing clipped vehicle-to-microphone distance, which is predicted from the mel-spectrogram of input audio, in a supervised learning approach. In addition, mel-spectrogram-based features are used directly for vehicle speed estimation, without introducing any intermediate features. The results show that the proposed features can be used for accurate vehicle detection and speed estimation, with an average error of $ 7.87 $ km/h. If we formulate speed estimation as a classification problem, with a $ 10 $ km/h discretization interval, the proposed method attains the average accuracy of $ 48.7\% $ for correct class prediction and $ 91.0\% $ when an offset of one class is allowed. The proposed method is evaluated on a dataset of 304 urban-environment on-field recordings of ten different vehicles.

\end{abstract}

%%%%%%%%%%%%%%%%%%%%%%%%%%%%%%%%%%%%%%%%%%%%%%%%%%%%%%%%%%%%%%%%%%%%%%%%%%%%%%%%

\section{Introduction}

Reliable, constant, and automatic traffic monitoring (TM) is important for adequate traffic law enforcement in most countries, and is considered an important tool in preventing road accidents and reducing fatalities. TM systems can benefit more from a wide range of traffic data to improve the performance of the roadway systems, enable adequate implementation of traffic law enforcement, prediction of future transport needs, and improve traffic safety. Traffic data usually include estimates of vehicle count, traffic volume, vehicle acceleration and speed, vehicle length, weight, and type \cite{Won}. They can also be used for detection of traffic irregularities and road accidents.

Current TM systems use different sensing technologies which can be divided into in-roadway-based (induction loop, piezoelectric sensor), over-roadway-based (infrared sensor, camera) and side-roadway-based (acoustic sensor, LIDAR) \cite{Won}. Lately, deep learning methods have been successfully deployed in vision-based TM systems, especially in vehicle tracking, vehicle identification and traffic anomaly detection \cite{Naphade}. Although vision-based TM systems operate well, they are complex, expensive, dependent on environmental conditions (reduced light intensity, shadows, vehicle headlights, etc.), and thus have limited application \cite{Won,Morris}.

Acoustic TM represents a viable alternative to the existing monitoring technologies and provides complementary information to visual surveillance systems. Acoustic-based TM has several advantages over other monitoring technologies. For example, with respect to cameras, microphones are cheaper, have lower power consumption, and require less storage space. They are not affected by visual occlusions and deteriorating ambient light conditions. They are easier to install and maintain, with low wear and tear. Acoustic sensors are less disturbing to drivers' behavior and have fewer privacy issues \cite{Won}, \cite{Wilson}.

The existing acoustic TM approaches are based on measurements with one microphone \cite{Quinn, Couvreur, Cevher, Barnwal, Kubera, Koops, Giraldo, Goksu} and microphone arrays \cite{Lopez, Lo, Marmaroli}. A more detailed overview of these approaches can be found in \cite{Djukanovic3}.

This paper deals with acoustic vehicle detection and speed estimation using single sensor measurements. We propose a supervised-learning method based on the short-time power spectrum of input audio. Vehicle detection is carried out by minimizing the clipped vehicle-to-microphone distance (CVMD), an approach introduced in \cite{Djukanovic1}. The position of CVMD minimum represents the closest point of approach (CPA) of a vehicle with respect to the microphone. The CPA instant is then used to localize a part of the short-time power spectrum of audio that will represent input features for speed estimation. The following short-time power spectrum representations are considered: (1) Mel spectrogram (MS), (2) Log-mel spectrogram (LMS), and (3) Log-mel cepstral spectrogram (MFCC). The proposed vehicle detection and speed estimation methods are trained and tested on a dataset of 304 on-field vehicle recordings \cite{Djukanovic3}.

In this paper, we improve the vehicle detection accuracy compared with \cite{Djukanovic3}. Experimental results show that two-stage neural network-based CVMD regression yields better results compared to its one-stage counterpart, which also holds for the vehicle counting task \cite{Djukanovic2}. We also show that the short-time power spectrum audio representations (MS, LMS and MFCC) can be used for reliable speed estimation, with MS yielding the lowest error of $ 7.87 $ km/h.

\pubidadjcol  %% place this in second column of the first page

%%%%%%%%%%%%%%%%%%%%%%%%%%%%%%%%%%%%%%%%%%%%%%%%%%%%%%%%%%%%%%%%%%%%%%%%%%%%%%%%

\section{Proposed vehicle detection and speed estimation}

Our supervised learning approach uses single microphone audio data (Section II-A). We propose to detect the pass-by instant of vehicle by minimizing its CVMD \cite{Djukanovic1}, \cite{Djukanovic2} (Section II-B). Then, we propose a speed estimation method and describe the corresponding input features (MS, LMS and MFCC) in Section II-C.

\begin{figure}[thpb]
\centering
\includegraphics[scale=0.33]{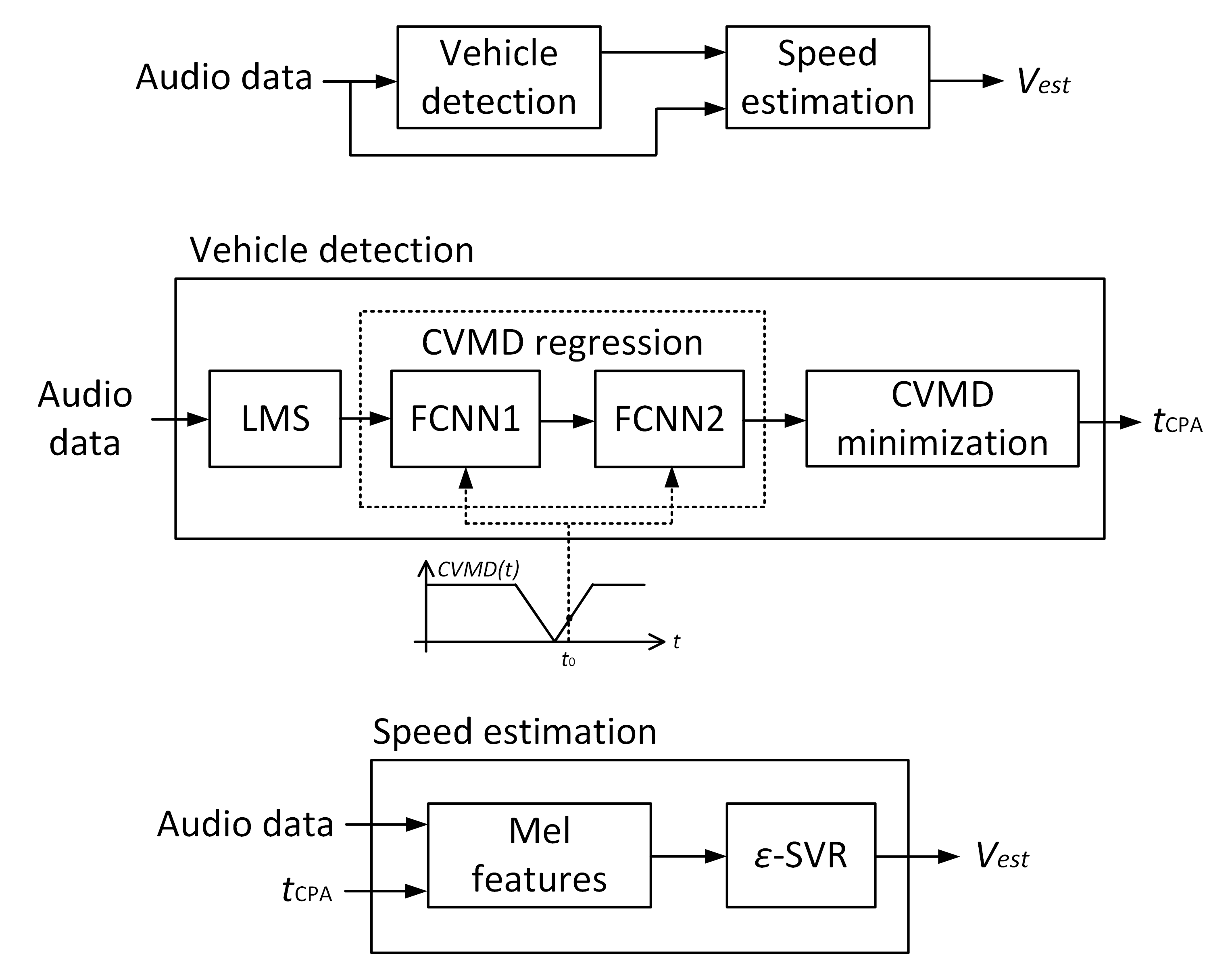}
\caption{{\it{Top:}} Block diagram of the proposed vehicle detection and speed estimation method. {\it{Middle:}} Vehicle detection is formulated as $ t_{\text{CPA}} $ estimation and it is implemented in a supervised learning fashion with two fully-connected neural networks (FCNNs) and CVMD minimization. {\it{Bottom:}} Speed estimation using mel-spectrogram-based features (MS, LMS and MFCC).}
\label{figBlockDiagram}
\end{figure}

The block diagram of our method is given in Fig. \ref{figBlockDiagram}.

%%%%%%%%%%%%%%%%%%%%%%%%%%%%%%%%%%%%

\subsection{Dataset}
\label{Dataset}

The dataset of on-field single-vehicle recordings from \cite{Djukanovic3} is used for training, validating and testing the proposed method. The data were recorded by a GoPro Hero5 Session camera at a local road in Podgorica, Montenegro. It contains 304 audio-video recordings of 10 different vehicles with constant speeds ranging from 30 km/h to 105 km/h. Each recording contains a single-vehicle pass-by. Additional 71 environmental noise recordings (no vehicles passing by) were collected to improve vehicle detection and speed estimation. Ten second-long audio sequences with 44100 Hz sampling rate, WAV format, and 32-bit float PCM, were extracted from the original audio-video recordings.

Annotation data contain the speed and pass-by time of the considered vehicles. Precise pass-by time is obtained by visual identification of a video frame in which the vehicle starts to exit the camera view, which approximately corresponds to the CPA.

%%%%%%%%%%%%%%%%%%%%%%%%%%%%%%%%%%%%

\subsection{Vehicle Detection}
\label{VehicleDetection}

We define vehicle detection as estimation of the CPA instant. To that end, we introduce CVMD as a form of distance between the vehicle and the microphone \cite{Djukanovic1}, \cite{Djukanovic2}:
\begin{equation}
	\label{eq:distance}
	d(t)=
	\begin{cases} 
		\left|t-t_{\text{CPA}}\right|, & \quad \left|t-t_{\text{CPA}}\right| < T_D\\
		T_D, & \quad  \text{elsewhere}, 
	\end{cases}
\end{equation}
where $ t_{\text{CPA}} $ is the vehicle's CPA instant and $ T_D $ represents a constant distance threshold for a vehicle too far from the microphone. We formulate CVMD estimation as a regression problem, that is, we estimate it using fully-connected neural networks (FCNNs) with LMS of audio as input features.

The proposed distance regression is presented in Fig. \ref{figBlockDiagram} (middle). Firstly, the LMS representation is calculated from the input audio signal. CVMD regression is then performed with FCNN1, having as inputs the LMS features, as proposed in \cite{Djukanovic2}. At each instant $ t $, the CVMD value is predicted using the LMS features within a narrow time interval centered at $ t $. FCNN2 serves to refine the output of FCNN1. To that end, FCNN2 takes as input a vector of successive intermediate CVMD predictions, centered at instant $ t $, and produces a refined CVMD prediction at $ t $. Finally, $ t_{\text{CPA}} $ is estimated by minimizing the predicted CVMD.

%%%%%%%%%%%%%%%%%%%%%%%%%%%%%%%%%%%%

\subsection{Speed Estimation}
\label{SpeedEstimation}

\begin{figure}[th!]
\centering
\includegraphics[scale=1]{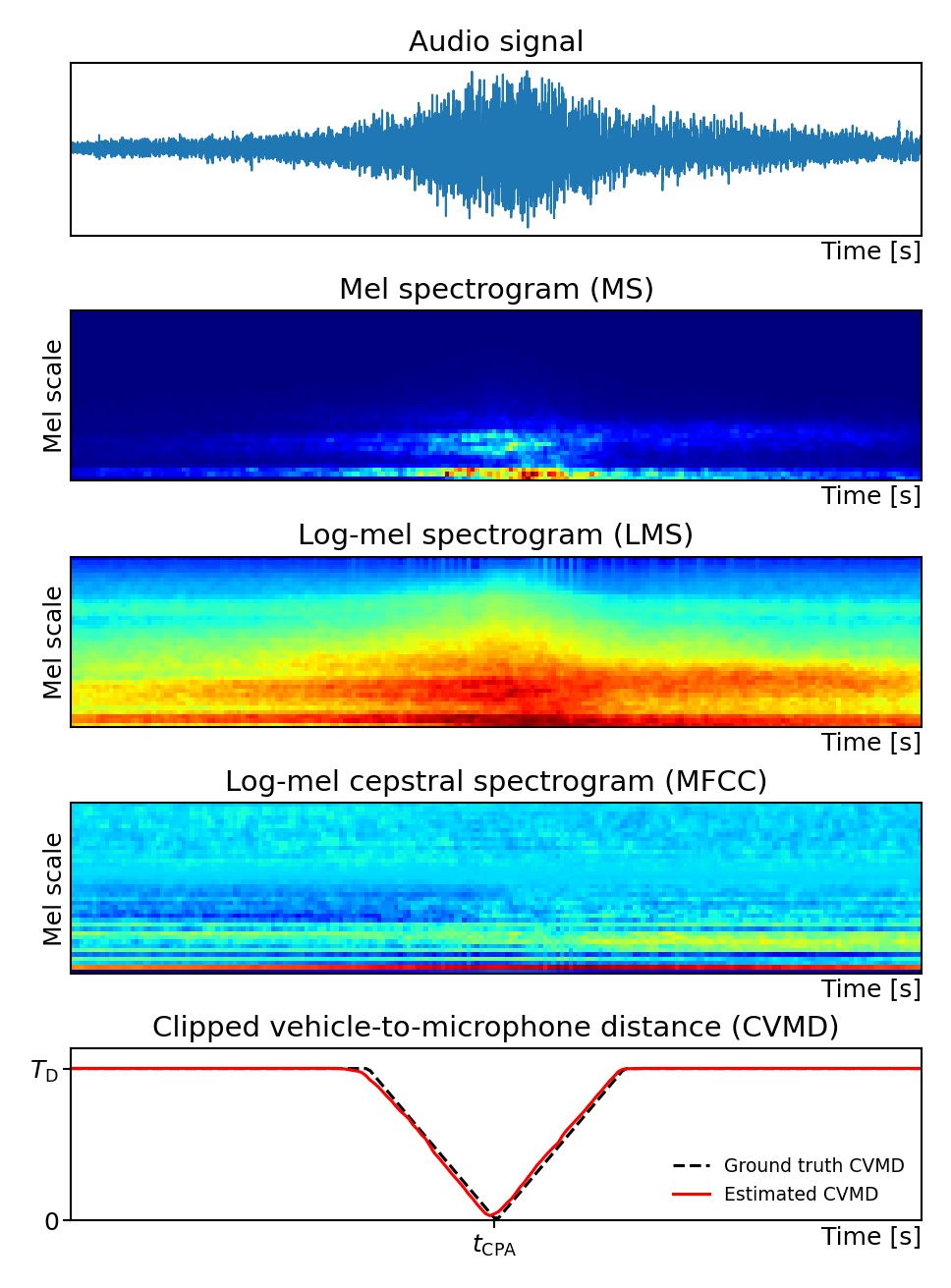}
\caption{{\it{First plot:}} Original audio signal. {\it{Second plot:}} Mel spectrogram of the audio signal. {\it{Third plot:}} Log-mel spectrogram of the audio signal. {\it{Fourth plot:}} Log-mel cepstral spectrogram of the audio signal. {\it{Fifth plot:}} Ground truth and predicted CVMD. The audio signal is clipped to 5 seconds and centered at $ t_{\text{CPA}} $.}
\label{figRepresentations}
\end{figure}

Mel-based audio representations are well suited as features in audio classification applications \cite{Serizel}. Promising results in speed estimation, obtained in \cite{Djukanovic3}, motivated us to further explore the potential of three mel-based audio representation variants, namely MS, LMS and MFCC. We analyze the impact when those are used directly as input features, without an intermediate speed-attenuation feature, proposed in \cite{Djukanovic3}. The MS represents the short-time power spectrum which remaps the original signal's frequency to the mel scale (logarithmic transformation of frequency bands). LMS is obtained as a logarithm of the MS magnitude. After the discrete cosine transform is applied to LMS, we get the MFCC representation. These features are presented in Fig. \ref{figRepresentations}.

Vehicle speed estimation (see Fig. \ref{figBlockDiagram} bottom) is performed using the $ \varepsilon $-support vector regression ($ \varepsilon $-SVR) \cite{Chang}. Only the MS, LMS and MFCC coefficients, around the estimated $ t_{\text{CPA}} $, are used as input features for speed estimation in the $ \varepsilon $-SVR block. Since the considered dataset is relatively small, $ \varepsilon $-SVR is selected as a speed estimation method over other approaches, such as neural networks. The output of the $ \varepsilon $-SVR block is the estimated speed.

%%%%%%%%%%%%%%%%%%%%%%%%%%%%%%%%%%%%

\subsection{Implementation details}

\subsubsection{Vehicle detection}

MS, LMS and MFCC audio representations are based on the short-time Fourier transform (STFT). The STFT block is implemented with the Hamming window of $ N_w=4096 $ samples ($ \approx  0.093$ s) and the hop length of $ N_h=0.27N_w=1105 $ samples ($ \approx  0.025$ s) \cite{Stankovic}. The $10$-second long audio files are sampled at $ 44100 $ Hz, which results in $ 400 $ STFT time frames, per file. The spectral resolution of all three mel representations is $ N_{mel}=40 $ mel bands, within the frequency range $ [0,16 \text{ kHz}] $.

The CVMD distance threshold is set to $ T_\text{D}=0.75 $ s, as in \cite{Djukanovic1}. CVMD regression is carried out using LMS as input features, with each feature vector containing the LMS coefficients from the current and $ Q=12 $ preceding and following time frames, with a stride of $ 3 $. The input features dimensionality is $M=(2Q+1)N_{mel}=1000$. The two FCNN configurations are set to $ 1000 $--$ 64 $--$ 64 $--$ 1 $ and $ 31 $--$ 31 $--$ 15 $--$ 1 $ neurons per layer, respectively. Both FCNNs use mean squared error loss, ReLU activation (linear activation in the last layer), $ L2 $ kernel regularization with factors $ 10^{-4} $ and $ 5\times 10^{-6} $, and $ 100 $ training epochs.

\subsubsection{Speed estimation}

We carried out grid search to extract the optimal $ \varepsilon $-SVR parameters $ C=150 $ (penalty of the error term) and $ \varepsilon=0.1 $ ($ \varepsilon $ determines the accuracy level of the approximated function). Additional two-dimensional grid searches gave the optimal mel-features' time and frequency window lengths. The following are the optimal time window lengths, centered at the estimated $ t_{\text{CPA}} $: $ {N^{t}_{\text{MS}}} = 91 $, $ {N^{t}_{\text{LMS}}} = 91 $, and $ {N^{t}_{\text{MFCC}}} = 61 $. The optimal mel-frequency windows are presented in the form of range of selected mel-frequency indices (from low to high): $ {N^{f}_{\text{MS}}} = [3,31] $, $ {N^{f}_{\text{LMS}}} = [2,20] $, and $ {N^{f}_{\text{MFCC}}} = [1,31] $.

We carry out $ 10 $-fold cross-validation in vehicle detection and speed estimation methods. One fold (vehicle) is used as a test and the remaining nine folds are used to train and validate the model. The cross-validation is iterated $ 10 $ times. The same train-validation split ($80\%$-$20\%$) is used in both methods, as described in \cite{Djukanovic3}.

%%%%%%%%%%%%%%%%%%%%%%%%%%%%%%%%%%%%%%%%%%%%%%%%%%%%%%%%%%%%%%%%%%%%%%%%%%%%%%%%

\addtolength{\textheight}{-0.9cm}

%%%%%%%%%%%%%%%%%%%%%%%%%%%%%%%%%%%%%%%%%%%%%%%%%%%%%%%%%%%%%%%%%%%%%%%%%%%%%%%%

\section{Experimental results}

\begin{figure}[t!]
	\centering
	\includegraphics[scale=0.81]{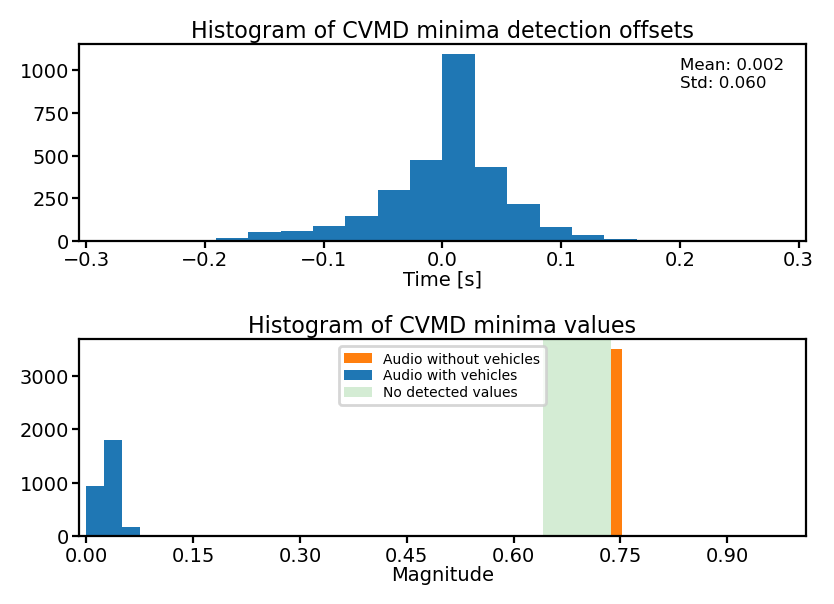}
	\caption{{\it{Top:}} Histogram of CVMD minima detection offsets. {\it{Bottom:}} Histogram of CVMD minima values. Green rectangle separates the vehicle (blue) and no-vehicle (orange) histograms.}
	\label{figCVMDhistograms}
\end{figure}

Vehicle detection error is evaluated on test data and calculated as offset between the true and predicted CVMD minima positions. Detection error histogram is presented in Fig. \ref{figCVMDhistograms} (top), with all $ 10 $ iterations included. We can model the detection error as a normal random variable with the mean value of $ 0.002 $ and standard deviation value of $ 0.06 $. The detection error is improved compared to \cite{Djukanovic3}, where reported values are $ -0.016 $ and $ 0.065 $, respectively. We can conclude that the proposed method is able to accurately detect the vehicle's CPA instant.

Vehicle detection accuracy is additionally evaluated in Fig. \ref{figCVMDhistograms} (bottom), where we compared the predicted CVMD minima histograms in the cases of test data with vehicles (blue histogram) and without vehicles (orange histogram) passing by. This is the reason additional $ 36 $ train and $ 35 $ test no-vehicle recordings were included in the experiment (Section \ref{Dataset}). The CVMD magnitude threshold should be set within the green rectangle separating the vehicle and no-vehicle histograms. The separating rectangle is much wider than in \cite{Djukanovic3}, so the vehicle appearance is better discerned with respect to no-vehicle cases.

Vehicle detection was also tested in a scenario with a one-stage FCNN setup in the CVMD regression. The obtained results showed that a two-stage setup is significantly better in predicting the CPA instant than its one-stage counterpart.

% \begin{figure}[thpb]

The root-mean-square error (RMSE) is used to evaluate the speed estimation performance 
\begin{equation}\label{RMSE}
	\text{RMSE}=\sqrt{\frac{1}{L}\sum\nolimits_{l=1}^{L}(v_{l}^{est}-v_{l}^{true})^2},
\end{equation}
where $ v_{l}^{est}$ and $v_{l}^{true} $ represent the estimated and true speed of the $ l $-th measurement (audio file), and $ L $ is the number of measurements. Speed estimation RMSE values per vehicle are shown in Table \ref{Tab1}. The impact of strong environmental noise resulted in Renault Scenic and Mazda 3 Skyactive estimation errors being notably high. On the other hand, speed estimations of Nissan Qashqai and Opel Insignia are very accurate.

\begin{table}[h]
	\centering
	\caption{Speed estimation RMSE}
	\begin{tabular}{ l r r r }
		\hline\hline
		\multirow{2}{4em}{Vehicle} & \multicolumn{3}{c}{RMSE [km/h]} \\
		 & \multicolumn{1}{c}{MS} & \multicolumn{1}{c}{LMS} & \multicolumn{1}{c}{MFCC} \\ \hline
		Citroen C4 Picasso & $ 4.00 $ & $ 4.77 $ & $ 7.66 $ \\ \hline
		Mazda 3 Skyactive & $ 11.51 $ & $ 15.16 $ & $ 15.66 $ \\ \hline
		Mercedes AMG 550 & $ 8.86 $ & $ 9.25 $ & $ 10.78 $ \\ \hline
		Nissan Qashqai & $ 4.38 $ & $ 5.81 $ & $ 7.44 $ \\ \hline
		Opel Insignia & $ 5.92 $ & $ 5.43 $ & $ 6.46 $ \\ \hline
		Peugeot 3008 & $ 9.15 $ & $ 7.99 $ & $ 7.93 $ \\ \hline
		Peugeot 307 & $ 7.87 $ & $ 8.11 $ & $ 10.37 $ \\ \hline
		Renault Captur & $ 6.61 $ & $ 7.26 $ & $ 8.29 $ \\ \hline
		Renault Scenic & $ 14.47 $ & $ 15.20 $ & $ 14.63 $ \\ \hline
		VW Passat B7 & $ 5.89 $ & $ 6.65 $ & $ 10.17 $ \\ \Xhline{1pt}
		Average & $ 7.87 $ & $ 8.56 $ & $ 9.94 $ \\ \hline
		\hline
	\end{tabular}
	\label{Tab1}
\end{table}

For the second evaluation metric, speed interval $ [30, 105] $ km/h is discretized with a step of $ 10 $ km/h, starting from $ 25 $ km/h. Vehicle sounds are classified into eight speed classes. This metric shows the probability of predicting the speed class that is $ \Delta $ classes distant from the true class. Table \ref{Tab2} presents the speed classification accuracy for $|\Delta|\leq 1$, when MS, LMS and MFCC audio features are used.

\begin{table}[h]
	\centering
	\caption{Speed class prediction probability}
	\begin{tabular}{ l r r r | r r r }
		\hline\hline
		\multirow{2}{4em}{Vehicle} & \multicolumn{3}{c|}{$ \Delta=0 $} & \multicolumn{3}{c}{$ |\Delta|\leq 1 $} \\ 
		& \multicolumn{1}{c}{MS} & \multicolumn{1}{c}{LMS} & \multicolumn{1}{c|}{MFCC} & \multicolumn{1}{c}{MS} & \multicolumn{1}{c}{LMS} & \multicolumn{1}{c}{MFCC} \\ \hline
		Citroen C4 Picasso & $ 83.5\% $ & $ 75.7\% $ & $ 57.8\% $ & $ 100.0\% $ & $ 99.1\% $ & $ 91.7\% $ \\ \hline
		Mazda 3 Skyactive & $ 49.7\% $ & $ 21.6\% $ & $ 12.5\% $ & $ 76.9\% $ & $ 63.1\% $ & $ 63.8\% $ \\ \hline
		Mercedes AMG 550 & $ 36.7\% $ & $ 40.0\% $ & $ 28.0\% $ & $ 94.3\% $ & $ 92.7\% $ & $ 86.7\% $ \\ \hline
		Nissan Qashqai & $ 62.8\% $ & $ 43.4\% $ & $ 40.7\% $ & $ 100.0\% $ & $ 97.2\% $ & $ 89.7\% $ \\ \hline
		Opel Insignia & $ 44.4\% $ & $ 58.9\% $ & $ 53.0\% $ & $ 99.3\% $ & $ 100.0\% $ & $ 99.6\% $ \\ \hline
		Peugeot 307 & $ 32.8\% $ & $ 31.4\% $ & $ 37.2\% $ & $ 95.9\% $ & $ 99.0\% $ & $ 95.5\% $ \\ \hline
		Peugeot 3008 & $ 30.6\% $ & $ 51.3\% $ & $ 36.8\% $ & $ 88.7\% $ & $ 85.8\% $ & $ 82.3\% $ \\ \hline
		Renault Captur & $ 55.8\% $ & $ 54.5\% $ & $ 48.2\% $ & $ 96.1\% $ & $ 91.8\% $ & $ 91.2\% $ \\ \hline
		Renault Scenic & $ 29.4\% $ & $ 30.9\% $ & $ 16.3\% $ & $ 61.7\% $ & $ 60.9\% $ & $ 61.4\% $ \\ \hline
		VW Passat B7 & $ 61.1\% $ & $ 46.3\% $ & $ 22.6\% $ & $ 97.1\% $ & $ 100.0\% $ & $ 92.3\% $ \\ \Xhline{1pt}
		Average & $ 48.7\% $ & $ 45.4\% $ & $ 35.3\% $ & $91.0\% $ & $ 89.0\% $ & $ 85.4\% $ \\
		\hline\hline
	\end{tabular}
	\label{Tab2}
\end{table}

Regarding mel-based audio representations impact, MS is the most reliable speed estimation feature, with the best average RMSE value of $ 7.87 $ km/h and the best average classification accuracies of $ 48.7\% $ ($ \Delta=0 $) and $ 91.0\% $ ($ |\Delta|\leq 1 $). However, LMS is also considered as an important classification feature, since it provides average classification accuracies very close to the MS-based ones. The presented results qualify mel-based features as an important component in our future vehicle detection and speed estimation research.

%%%%%%%%%%%%%%%%%%%%%%%%%%%%%%%%%%%%%%%%%%%%%%%%%%%%%%%%%%%%%%%%%%%%%%%%%%%%%%%%

\section{Conclusions}

This paper explores the potential of using the mel-spectrogram features in vehicle speed estimation. The experimental results show that carefully selected mel features can be used directly in speed estimation, without intermediate and hand-crafted features.

In order to improve the performance of acoustic speed estimation, our future research will focus on data-oriented approaches. We will consider the application of data augmentation methods and the existing dataset will be extended with additional vehicles. Acoustic features and their modifications will be furtherly analyzed to improve the estimation accuracy.

%%%%%%%%%%%%%%%%%%%%%%%%%%%%%%%%%%%%%%%%%%%%%%%%%%%%%%%%%%%%%%%%%%%%%%%%%%%%%%%%

\bibliographystyle{IEEEbib}
\bibliography{paper}

\end{document}